%% file: root.tex
\documentclass[twoside,11pt]{article}

\usepackage{jmlr2e}
\usepackage{amsmath}
\usepackage{amsfonts,amssymb}
\usepackage{graphicx}        %
\usepackage{algorithm}
\usepackage[noend]{algpseudocode}
\usepackage{fixltx2e}
\usepackage{multicol}
\usepackage{booktabs}
\usepackage{subfig}
\usepackage[font=scriptsize]{caption}
\usepackage{comment}
\usepackage{epstopdf}
\usepackage{outlines}
\usepackage{multirow}
\usepackage{textcomp}
\usepackage{listings}
\usepackage[normalem]{ulem}
\usepackage{url}
\usepackage{mathtools}
\usepackage{svg}
\usepackage{color,soul}

\usepackage{hyperref}
\hypersetup{colorlinks = true,allcolors=blue}

\usepackage{enumitem}
\setitemize{noitemsep,topsep=1pt,parsep=1pt,partopsep=1pt}

\ShortHeadings{AirSim Drone Racing Lab}{Madaan et al.}
\firstpageno{1}

\begin{document}
\title{AirSim Drone Racing Lab}
\author{\name Ratnesh Madaan$^{1}$ \email ratnesh.madaan@microsoft.com \\
    \name Nicholas Gyde$^{1}$ \email v-nigyde@microsoft.com \\
    \name Sai Vemprala$^{1}$ \email sai.vemprala@microsoft.com \\
    \name Matthew Brown$^{1}$ \email v-mattbr@microsoft.com \\
    \name Keiko Nagami$^{2}$ \email knagami@stanford.edu \\
    \name Tim Taubner$^{2,3}$ \email taubnert@inf.ethz.ch \\
    \name Eric Cristofalo$^{2}$ \email ecristof@stanford.edu\\
    \name Davide Scaramuzza$^{3}$ \email sdavide@ifi.uzh.ch \\
    \name Mac Schwager$^{2}$ \email schwager@stanford.edu \\
    \name Ashish Kapoor$^{1}$ \email akapoor@microsoft.com \\
    \addr $^{1}$ Microsoft, One Microsoft Way, Redmond, WA 98052\\
    \addr $^{2}$ Multi-Robot Systems Lab, 
Department of Aeronautics and Astronautics, Stanford University \\
Durand Building, 496 Lomita Mall, Stanford, CA 94305\\
    \addr $^{3}$ Robotics and Perception Group, Depts. of Informatics and Neuroinformatics, University of Zurich and ETH, Andreasstrasse 15, 8050, Zurich, Switzerland
}
\maketitle
\vspace{-10mm}
\input{inputs/abstract}
\begin{keywords}
Drone Racing, UAV, Robotics, Planning, Perception, Machine Learning
\end{keywords}

\input{inputs/intro}

\input{inputs/related}
\input{inputs/airsim_drone_racing_lab}
\input{inputs/baselines}

\input{inputs/game_of_drones}

\input{inputs/winning_teams}

\input{inputs/conclusion_and_future}
\input{inputs/appendix}

\vskip 0.2in
\bibliography{references}

\end{document}

%% file: inputs/abstract.tex
\begin{abstract}%
Autonomous drone racing is a challenging research problem at the intersection of computer vision, planning, state estimation, and control. 
We introduce AirSim Drone Racing Lab, a simulation framework for enabling fast prototyping of algorithms for autonomy and enabling machine learning research in this domain, with the goal of reducing the time, money, and risks associated with field robotics. 
Our framework enables generation of racing tracks in multiple photo-realistic environments, orchestration of drone races, comes with a suite of gate assets, allows for multiple sensor modalities (monocular, depth, neuromorphic events, optical flow), different camera models, and benchmarking of planning, control, computer vision, and learning-based algorithms. 
We used our framework to host a simulation based drone racing competition at NeurIPS 2019.
The competition binaries are available at our github repository\footnote{\url{https://github.com/microsoft/AirSim-NeurIPS2019-Drone-Racing}}. 
\end{abstract}

%% file: inputs/intro.tex
\vspace{-4mm}
\section{Introduction}\label{intro}

Machine Learning (ML) methods are increasingly showing promise in addressing challenges in robotics and autonomous systems. 
Perception-action loops are at the core of these devices, and recent advances in reinforcement learning (RL) and imitation learning (IL) are potentially applicable in this domain. 
However, these results they are often limited to constrained laboratory settings~\citep{hwangbo2017control, kahn2017uncertainty} or focus on simple computer games~\citep{mnih2013playing, silver2016mastering}. 
ML has the potential to positively accelerate the field of robotics; however, the 
barrier to entry in this research area is high due to the financial cost, complexity and risk associated with having robots operating in the real world. 
In this paper we address the challenge of lowering the barrier to entry for ML researchers into the task of autonomous drone racing \citep{moon2019challenges}. 
Our core hypothesis is that simulation can help mitigate the complexity associated with experimenting with flying robots in the real world, and that it is possible to design APIs at an abstraction level targeted to ML researchers. 

Started by passionate hobbyists, first person view (FPV) drone racing has gained in popularity over the past few years and is now recognized as a professional sport.
Humans are able to fly drones (quadcopters in this context) aggressively 
through complex and cluttered environments, which may consist of varying number of gates, gate poses, sizes, texture and shapes, using only low resolution, noisy first-person view images while competing against and avoiding other racing drones in flight. 
The control inputs for the racing drones are angular velocity rates and thrust commands, which require the pilot to constantly stabilize them, unlike velocity or position input modes found on their commercial counterparts. 

The UAV research community has recognized drone racing as one of the next important challenges to address because it requires multiple autonomy modules (state estimation, gate detection and pose estimation, trajectory planning, control, and reactive avoidance strategies) to work in harmony with each other.
However, quick prototyping of the autonomy stack and verifying its generalization abilities on varied race environments is still a challenge. 

\begin{figure*}
    \centering
    \includegraphics[width=\textwidth]{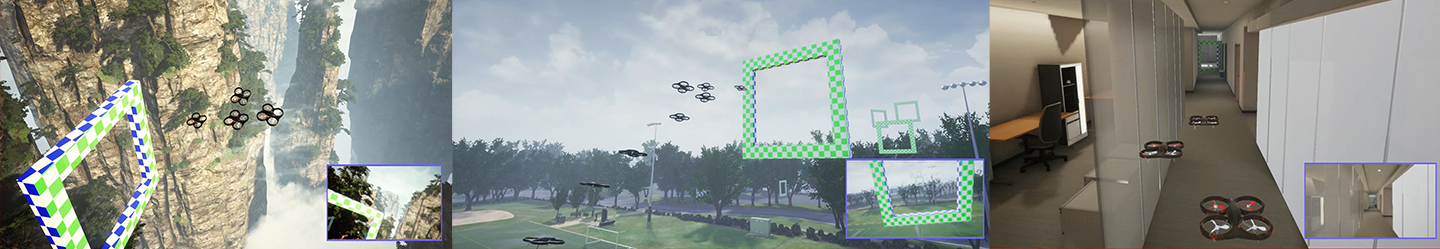}
    \caption{Snapshots from a few of our racing environments and tracks.}
    \label{fig:cover_montage}
\end{figure*}

We develop a simulation framework, AirSim Drone Racing Lab, to address the above. 
In addition, we used our framework to organize a simulation based drone racing competition at the NeurIPS 2019 Conference, \textit{Game of Drones (GoD)}, with the goal of attracting the ML community to focus on robotic sub-problems of trajectory planning, control, computer vision, and head-to-head competition with an opponent racer.

%% file: inputs/related.tex
\vspace{-4mm}
\section{Related Work}
\vspace{-2mm}
The field of robotics has a rich history of challenges and competitions being drivers of research.
In recent years, drone-specific programs such as~\cite{darpafla}, \cite{mbzirc}, and~\cite{imav} have come into focus, where the objective has generally been to build UAV platforms to address tasks such as navigation, mapping, and pick-and-place autonomously. 

Autonomous drone racing has gained substantial momentum over the last few years as a research problem, 
with multiple approaches being presented to address perception, planning and control to achieve agile, accurate flight~\citep{jung2018perception, kaufmann2018deep, kaufmann2019beauty, lin2019flying, loquercio2019deep}. 
Research oriented drone racing competitions have also been introduced, the largest among them being the recent~\cite{alphapilot}, which used the MIT FlightGoggles simulation framework~\citep{guerra2019flightgoggles} in its qualification round and required implementation on their hardware in the later stages. 
Similarly, an annual hardware oriented drone racing challenge~\citep{moon2019challenges} has been hosted at the IROS robotics conference since 2016. 
In addition, multiple datasets (UZH-FPV~\citep{delmerico2019we}, Blackbird~\citep{antonini2018blackbird}) have also been released to aid the development of vision based drone racing algorithms. 

The goal of our AirSim Drone Racing Lab simulation framework is to help bridge the gap between simulation and reality by utilizing high fidelity graphics and physics simulation. 
We achieve this by building on top of AirSim~\citep{shah2018airsim}, which leverages Unreal Engine for graphics. The graphical capabilities of Unreal Engine enable generation of photo-realistic environments, benefiting the development of perception algorithms and sim-to-real transfer techniques. For example,~\cite{loquercio2019deep} attempt sim-to-real transfer by randomizing the visual properties of the underlying Gazebo simulator~\citep{koenig2004design} and the RotorS extension~\citep{furrer2016rotors}. 
However, Gazebo does not quite match the photorealistic capabilities of Unreal Engine. 

Through the \textit{Game of Drones (GoD)} competition at NeurIPS 2019, we put the framework to test. 
While other drone racing competitions prioritized the design of/implementation on hardware racing drones, we believe simulation helps reach wider audiences and connect the machine learning and robotics communities, while allowing participants to focus on autonomy-enabling algorithms. Unlike the AlphaPilot and IROS racing challenges which primarily use a time trial format, the participants in \textit{GoD} raced against a simulated opponent, emphasizing head-to-head competition and bringing the problem statement closer to human FPV drone racing. 

%% file: inputs/airsim_drone_racing_lab.tex
\vspace{-4mm}
\section{AirSim Drone Racing Lab}
\vspace{-2mm}

One of the main goals of our framework is to make drone racing accessible to ML researchers and engineers, who have the relevant knowledge of algorithms and software but might not have exposure to the hardware and systems aspects of robotics. 
To achieve this, we build on AirSim~\citep{shah2018airsim}, a high-fidelity simulation framework for multirotors which implements a lightweight physics engine, flight controller, and inertial sensors; and also comes with photo-realistic camera and depth sensors via Unreal Engine (UE). 

\noindent While it provides us with a good starting point, drone racing research with AirSim is not possible out of the box. 
Drone racing is a research problem which involves multiple moving components. 
Apart from the core AirSim components, 
one needs a framework to orchestrate drone races, and a set of APIs and features geared towards ML research. 

\textit{Drone-racing specific needs} include multiple \textit{environments} in which tracks can be set up, a library of 3D assets corresponding to \textit{drone gates}, a framework to start and reset \textit{races}, and track the \textit{race progress} (every \textit{gate} has to detect which drone \textit{racer} passed through it), displaying and accessing \textit{race progress} from APIs to monitor score and performance, and an ability to enforce \textit{race rules} such as time \textit{penalties} associated with environment collisions and \textit{disqualifications} in the event of drone-drone collision. 

\begin{figure*}
	\centering
	\includegraphics[width=\textwidth ]{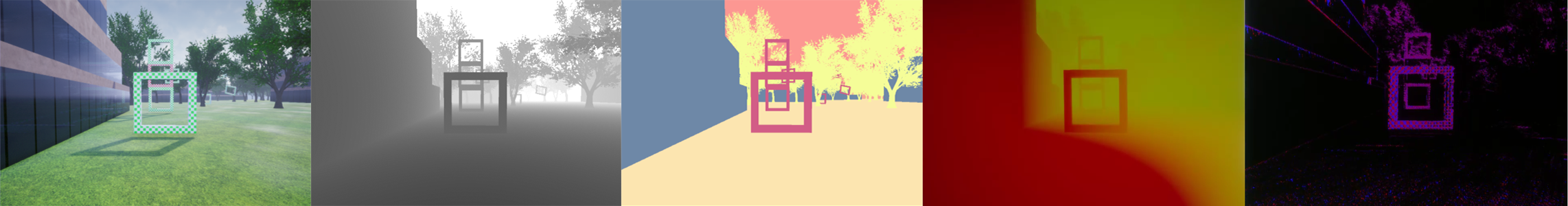}
	\caption{An example showcasing different sensor modalities our framework provides - from left to right: monocular RGB, depth, segmentation, optical flow, and events.}
	\label{fig:cameratypes}
\end{figure*}
\textit{ML specific needs} include data acquisition from multiple sensor modalities, generalization of learnt models (whether perception or control oriented); representation, imitation, and reinforcement learning.  
Specifically, we want our APIs to allow for generation of diverse gate datasets (spawning and destructing gate assets); domain randomization (changing the shape, scale, pose, and texture of gates), and cater for multiple components of autonomy used in drone racing targeted at different communities. 
To emphasize the last point, we provide high level planning and control APIs for researchers who are interested in solving perception tasks (detecting gates from images and estimating their pose), low-level controls for users interested in learning sensorimotor policies, and support for imitation learning dataset generation. 
In order to train policies robust to different camera models, we also provide APIs to set camera model parameters - intrinsics, distortion coefficients, and rolling shutter; while also generating ground truth optical flow images. 
\begin{figure*}
    \centering
    \includegraphics[width=1.0\textwidth ]{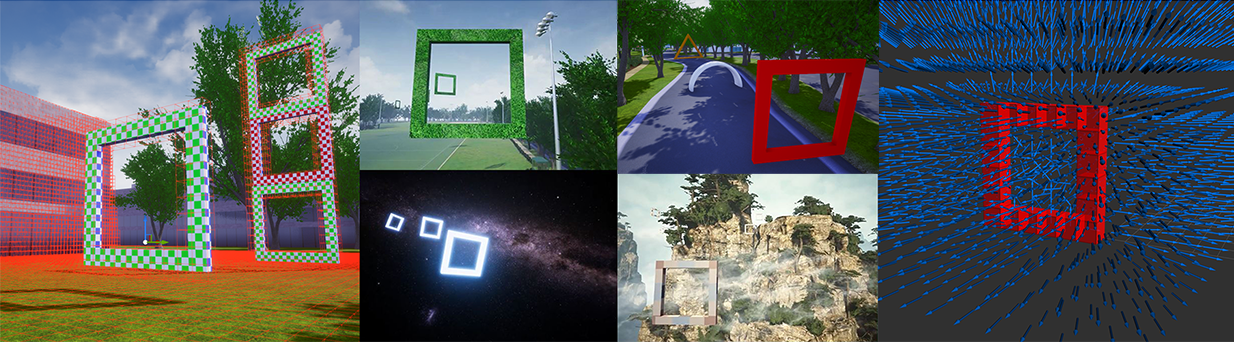}
    \caption{Sample features from AirSim Drone Racing Lab: Left - Environment ground truth via voxel grid, Right - Signed Distance Field gradient visualization for a gate, Middle - Examples of Domain Randomization.}
    \label{fig:domainrandomization}
\end{figure*}

In addition, high speed drone racing creates a need for investigating into event cameras due to their high frame rate and dynamic range, asynchronous nature, and lack of motion blur~\citep{Rebecq18corl, gallego2019event}. 
We leverage Unreal Engine's graphics capabilities to draw events at a high rate, and we aim to provide a full framework to output event data in the near future (Appendix \ref{appendix:eventsandopticalflow}). 
Finally, we also provide ground truth representations of the full simulated world or a region of interest thereof, in terms of voxel grids, specifically sparse voxel octrees~\citep{laine2010efficient}. 
These voxel grids are converted into signed distance fields~\citep{oleynikova2017voxblox, han2019fiesta}, that provide the signed distance to the nearest obstacle and the gradient thereof at query locations, which can then be consumed by trajectory optimization based planners. 

\noindent \textbf{Software Architecture:} A schematic of our software architecure is visualized in \autoref{fig:architecture}. 
Each race environment contains custom UE actors (\texttt{ADrone}, \texttt{AGate}) which detect race-related simulation events - drone collisions, crashes, near misses, and gate passthroughs. 
The actors then relay these events to a standalone C++ race module separated from AirSim, which aids in orchestration of a races. %
The race module is composed of multiple classes (\texttt{Race}, \texttt{Racer}, \texttt{Gate}), each of which have relevant UE actors (\texttt{ADrone}, \texttt{AGate}). 
Each \texttt{Racer} owns a \texttt{RacerProgress} object, which tracks its competition metrics based on events (for example, a collision event may be interpreted as a disqualification). 
Finally, the \texttt{CompManager} collects the overall race state from all the \texttt{RacerProgress} objects, and streams relevant telemetry to a log file. 
The file is flushed in real time and exists as a channel to inform the AirSim client of the race state in real-time, and serves as the final component in a continuous feedback loop between API control and \texttt{Race} state tracking. 
We note here that for \textit{GoD}, the participants submitted the generated log files to our server where we ran the evaluation scripts, and then updated the leaderboard.

In addition, we allow for changing of UE environments at runtime by cooking them into DLC packages (\texttt{.pak} files) and exposing an API. 
A full list of our APIs is available in Appendix \ref{appendix:api} and our website. 

%% file: inputs/baselines.tex
\vspace{-4mm}
\section{Race Tracks and Baselines}
\vspace{-2mm}
Out current release binaries include three Unreal Engine environments - \texttt{Soccer Field}, \texttt{ZhangJiaJie}, \texttt{MSR Building 99} - over which racing tracks are designed, with a cumulative of nine tracks. 
We used three tracks each in different stages (training, qualification, final round) of \textit{GoD}.
In order to benchmark autonomy algorithms, it is important to characterize the complexity of a racing track quantitatively. 
We introduce two measures of complexity: curvature per unit length (for planning and control tasks) and next gate visibility (for perception oriented tasks), and then outline baseline algorithms for the same. 

\begin{figure*}
    \centering
    \includegraphics[width=0.65\textwidth ]{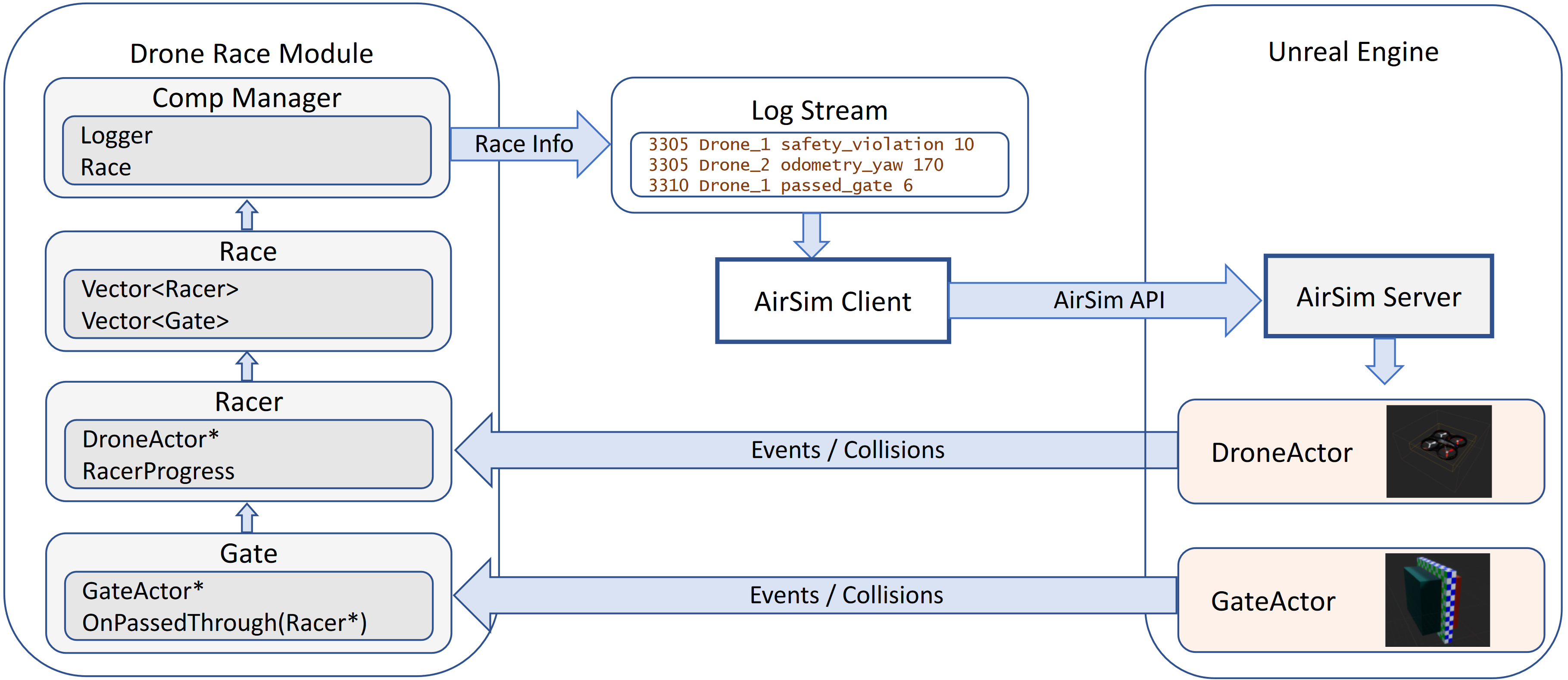}
    \caption{AirSim Drone Racing Lab software architecture.}
    \label{fig:architecture}
\end{figure*}

\textbf{Quantifying complexity of a race track:}
For \textit{planning and control tasks}, we use the curvature measure of a 3D curve and normalize it per unit length. A racing track can be defined by the number of gates, their poses, and the direction vector along which a racer should pass through the gate. We fit a 3rd order spline through the center of all gates. 
For this three dimensional curve, given by $\gamma(t) = (x(t), y(t), z(t))$, we calculate the instantaneous curvature as

\tiny{\begin{equation*}
    \kappa=\frac{\sqrt{\left(z^{\prime \prime} y^{\prime}-y^{\prime \prime} z^{\prime}\right)^{2}+\left(x^{\prime \prime} z^{\prime}-z^{\prime \prime} x^{\prime}\right)^{2}+\left(y^{\prime \prime} x^{\prime}-x^{\prime \prime} y^{\prime}\right)^{2}}}{\left(x^{\prime 2}+y^{\prime 2}+z^{\prime 2}\right)^{\frac{3}{2}}}
\end{equation*}}
\normalsize
We plot the instantaneous curvature against track length (\autoref{fig:tracks_curvature}), and define a scalar track \textit{Curvature Metric} quantifying its cumulative complexity -- AUC (Area Under the Curve) \textit{normalized} by the track length, and report it in Table \ref{tab:baseline_performance}.
For \textit{perception tasks}, we define hardness by \textit{Gate Visibility} -- the number of pixels belonging to the next gate in the RGB image divided by the total pixel area of the image (\autoref{fig:tracks_gate_visibility}).
This gives an approximate measure of the information available of the next gate in the FPV image when flying along the baseline trajectory. 
We note that this is only an indicative metric and actual gate visibility can vary with camera intrinsic parameters such as field of view. 
We refer the reader to Appendix \ref{appendix:trackcomplexity} for more details.

\textbf{Trajectory planning and tracking baseline:}
Researchers focusing on perception oriented tasks would not want to spend a lot of their resources on planning and control oriented tasks. In addition, it is important that the training datasets they use are a representative sample of test time, when a drone is flying through gates. 
To address these needs, we introduce a high level API \texttt{moveOnSpline}, which consumes a list of 3D waypoints along with desired velocity and acceleration limits, and does minimum jerk trajectory planning and pure pursuit tracking control in the backend.

Exploiting quadrotors' differential flatness property (\cite{mellinger2011minimum}), we plan in $\mathbb{R}^3$ for minimum jerk trajectories which are splines composed of piece-wise polynomials with {$C^{2}$} continuity. 
We plan for yaw angles separately and assign them to follow the tangents of the fitted spline. 
We refer the reader to~\citep{oleynikova2016continuous, burri2015real-time, richter2016polynomial} for details. 
To track the trajectory, we implement a pure-pursuit controller, which minimizes position and velocity errors in the cross-track, along-track, and world $z$ dimensions. 
We report the lap times in Table \ref{tab:baseline_performance}, along with the maximum specified velocity and acceleration for each track (for the \textit{Building 99} track, these values are much lower due to tight corridors). 

In the context of drone racing, it is useful to specify a user-defined velocity direction vector while passing a user-defined waypoint (for example -- flying along the gate normal direction through a gate center). 
For this use case, we expose a \texttt{moveOnSplineVelConstraints} API which takes in velocity vector constraints in addition to each 3D waypoint. 

\begin{table}[]
\centering{
\resizebox{0.9\textwidth}{!}{%
\begin{tabular}{@{}cccccc@{}}
\toprule
\textbf{Track Name} & \textbf{Curvature Metric ($m^{-1}$)} & \textbf{$v_{max}(m/s),\ a_{max}(m/s^2)$} & \textbf{\texttt{moveOnSpline} Lap Time ($s$)} & \textbf{Perception Error ($m$)}  \\ \midrule
T\_Soccer\_Field\_Easy  & 0.05 & 30, 15 & 58.019 &1.673 \\
T\_ZhangJiaJie\_Medium & 0.04 & 30, 15  & 125.775 &2.097 \\
T\_Building\_99\_Hard & 0.09 & 5, 2 & 80.699 & 1.539 \\
Q\_Tier\_1 & 0.04 & 30, 15 & 110.867 & 2.845 \\
Q\_Tier\_2 & 0.03 & 30, 15 & 80.553  & 1.640 \\
Q\_Tier\_3 & 0.02 & 30, 15 & 143.830 & 2.481 \\
F\_Tier\_1 & 0.05 & 30, 15 & 106.637 & 2.856 \\
F\_Tier\_2 & 0.03 & 30, 15 & 113.788 & 3.774 \\
F\_Tier\_3 & 0.02 & 30, 15 & 124.135 & 1.739 \\ \bottomrule
\end{tabular}}
\caption{Track complexity and baseline performance.} 
\label{tab:baseline_performance}
}
\end{table}

\textbf{Gate detection baseline:}
We use a recursive filtering algorithm to continuously estimate the relative pose of the next gate with respect to the drone. 
The gates used in the competition tracks are rectangular, planar, and have similar color and texture. Under the assumptions of known gate dimensions, we use planar homography to estimate the center of the gate (\cite{Hartley2004}). 
We capture a baseline image of a gate under the assumption of known gate dimensions and relative distance from the camera, with the camera's optical axis normal to the gate cross section. 
With this setup, the center pixel of the baseline image can be mapped to the gate center. 

During flight, we use color thresholding to extract a gate mask and retrieve the best estimates of the four gate corners. The homography matrix is determined to map the center gate pixel of the baseline image to that of a different image, using the four gate corner pixel points as a reference. 
From here, a point to point correspondence is used to find the 3D coordinate of the gate center. 
The measurements of the next gate's center location are fed into a Kalman Filter to obtain improved estimates. 

In order to quantify the performance of this baseline, we collect $1000$ measurements of gate centers, by flying the quadrotor through each track using \texttt{moveOnSpline} (assuming ground truth of gate positions), and report the mean Euclidean distance between gate center measurements and ground truth in \autoref{tab:baseline_performance}.

%% file: inputs/game_of_drones.tex
\vspace{-4mm}
\section{Game of Drones Competition at NeurIPS 2019}
We used our framework to host a simulation based drone racing competition at NeurIPS 2019.  
The competition was scoped to focus on three core areas pertinent to autonomous drone racing -  perception, trajectory planning and control, and head-tp-head competition with a \textit{single} competitor drone. 
(We note that our framework allows for an arbitrary number of drones in a single race, but \textit{GoD} was limited to two racers.) 
In each tier, the objective is to pass through all the gates in minimum possible time, without any collision with the environment. 
If an opponent racer is present in the tier, there is an additional objective to not collide with it. 
In the case of collision with the gates or any other objects in the environment, a collision penalty to the total lap time is applied.  
In the case of drone-drone collision, the trailing drone is disqualified. 
The participants are first ranked according to the number of gates flown through, followed by lap time. 
The competition specific binaries and baselines' implementations are available at our github repository\footnote{\url{https://github.com/microsoft/AirSim-NeurIPS2019-Drone-Racing}}. 
We now explain each tier, and a competition specific game-theoretic planning baseline.

\textit{Tier 1 -- Planning only:} 
The participant’s drone races t\^ete-\`a-t\^ete with an opponent racer. 
Ground truth for state estimation and environment is provided via our APIs, in the form of the odometry (position and velocity) of the participant and the opponent drones, and the poses of all the gates. 
The opponent racer follows a minimum jerk trajectory via \texttt{moveOnSpline}, and goes through randomized waypoints selected in each gate's cross section. 
Hence, the opponent's trajectory varies at every run. 

\begin{figure*}
    \centering
    \subfloat[Tier 1\label{fig:tier_1}]{\includegraphics[width=0.3\textwidth]{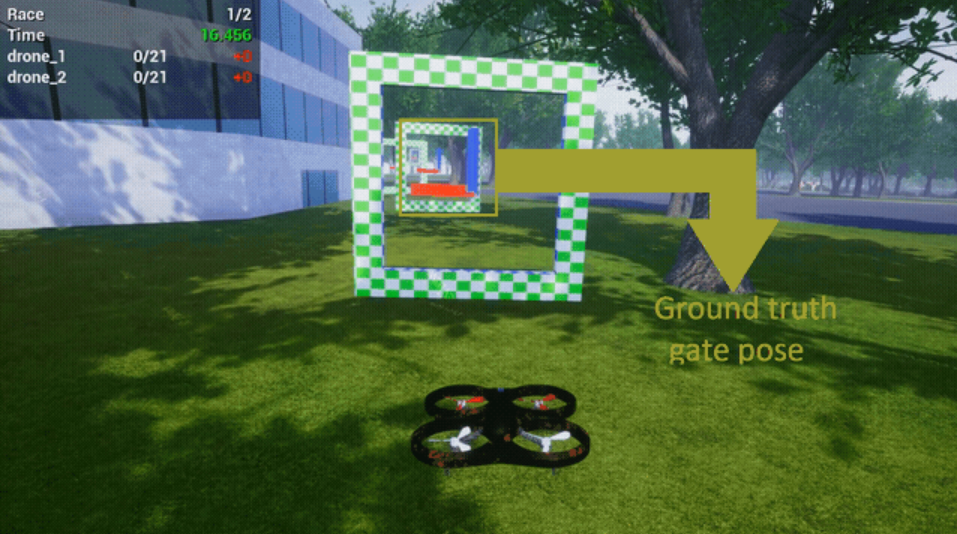}}
    \qquad
    \subfloat[Tier 2 and 3\label{fig:tier_2}]{\includegraphics[width=0.3\textwidth]{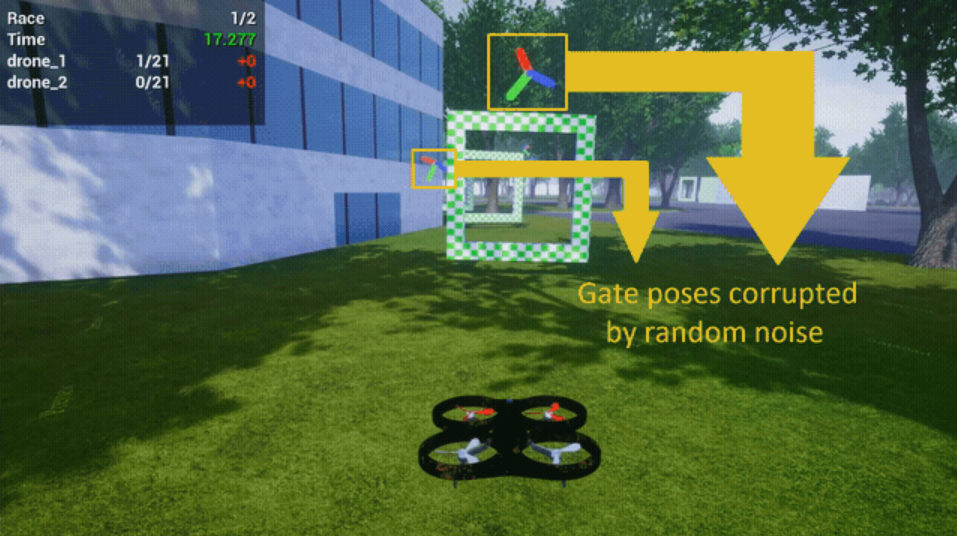}}
    \caption{Ground truth gate poses are given in Tier 1; poses are corrupted with noise in Tiers 2-3.}
    \label{fig:overview_of_tiers}
\end{figure*}

\textit{Tier 2 -- Perception only:} 
In this tier, the gate poses returned by the API are corrupted with noise as shown in \autoref{fig:tier_2}, and there is no opponent drone. 
The ground truth state estimate of the participant drone is still available. 
The next gate is not always in view, but the noisy pose returned by our API help steer the participants roughly in the right direction, after which vision-based control would be necessary.

\textit{Tier 3 -- Perception and Planning:}
This tier has both noisy gate poses and an opponent racer, and is essentially a combination of Tier 1 and 2. 

\textit{Game theoretic planning baseline:}
Head-to-head drone racing in Tiers 1 and 3 of \textit{GoD} brings with it an inherent competitor interaction which can be addressed via game theory. 
We implemented a Game Theoretic Planner baseline based on~(\cite{spica2018realtime}) which can solve the two-player drone racing problem.
We refer the reader to our github repository for the implementation of this baseline. 

\textbf{Organizational details:}
We ran \textit{GoD} in two stages: a qualification and a final round. 
Initially, A set of \textit{training} binaries with our three UE environments was released to the participants, for prototyping 
on user-defined race tracks and user-defined race tiers. 
Then, we released a new set of binaries for the  \textit{qualification} stage with one race track for each tiers, where participants were asked to make submissions for their choice of tiers. 
We saw 117 teams registering for the competition worldwide, with 16 unique entries that showed up on the \textit{qualification} leaderboard. 
Lastly, the \textit{final} round was conducted with another set of racetracks, again with one track for each tier.

%% file: inputs/winning_teams.tex
\textbf{Winning teams' approaches:}
The detailed reports, talks of the winning teams and the leaderboards can be found at our website\footnote{\url{https://microsoft.github.io/AirSim-NeurIPS2019-Drone-Racing}}. 
We briefly summarize their approaches here\footnote{We do note that this section is not our work, and the relevant publications by the winning teams should be cited if the reader wishes to build upon their work. 
However, we believe it is important to disseminate the core components of their approaches.}.

Tier 1 winner (\textit{Team Dedal\'e}) won by a significant margin because of two key components in their approach - (1) planning a globally optimal trajectory by allowing it to pass anywhere through the gate cross-sections (as opposed to forcing an artificial position constraint at the gate center), (2) modeling of drag forces, (3) a non-linear model predictive controller~\citep{kamel2017linear}.  
It is a known fact that the drag forces acting on any body (including a quadrotor) become significant at high speeds~\citep{spedicato2017minimum}, and that the actual velocity of the quadrotor does not match the reference velocity generated by methods which do not account for drag~\citep{richter2016polynomial, burri2015real-time}.
        
Tiers 2 and 3 winner (\textit{Team Sangyun}) also won by a significant margin. 
For gate detection, they use MobileNet-SSD~\citep{howard2017mobilenets,liu2016ssd} trained on a a synthetic dataset generated from our training binaries. 
The predicted gate bounding box is compressed into a vectorized state and fed into a policy network, which is trained in an actor-critic fashion with the reward function pushing the policy to fly through the center of the next gate. 
They had a head start on their RL policy by using pre-trained weights from their previous work on AirSim~\citep{shin2020reward}. 
When flying between gates, they use genetic algorithms~\citep{man1996genetic} for optimizing the inputs parameters to our \texttt{moveOnSpline} API, and suggested that this helped a lot in improving their lap times.   
Between gates, they fly to a weighted sum of the current position of the drone and the given noisy gate pose.

Tier 2 and 3 second position (\textit{Team USRG}) also used MobileNet-SSD for the perception problem, but in addition to detecting the closest gate, they also detect competitor drone. 
For control, they used a hybrid scheme which switches between position based control (when there is no gate detected) and a velocity based control (when next gate is visible). 
For position based control, they use the weighted sum strategy similar to \textit{Team Sangyun}, whereas the velocity based control is essentially a rule-based servoing method to fly through the center of the next gate~\citep{jung2018perception}.

%% file: inputs/conclusion_and_future.tex
\vspace{-4mm}
\section{Conclusion}
\vspace{-1mm}
We introduce a simulation framework targeted at the domain of autonomous drone racing, with the goal of reducing the entry barrier for the machine learning community and benchmarking of various modules of autonomy in a simulation environment. 
We build on Microsoft AirSim, and extend its capabilities by adding the ability to orchestrate drone races, exposing additional input modalities (e.g. event cameras and optical flow), customizing camera sensor models, providing environment ground truth, marker visualizations, high level trajectory planning and tracking features, and modifying flight controller gains. 
We used our framework to host a simulation-only drone racing competition at NeurIPS 2019, which saw a wide range of participation and corresponding approaches. 

%% file: inputs/appendix.tex
\newpage
\appendix

\section{AirSim Drone Racing API}
\label{appendix:api}
We highlight the main components of AirSim Drone Racing Lab API below, which augment the existing AirSim API functions.
Note: APIs marked with a `\#' superscript are not available in the GoD binaries, but we will release them along with our full framework:

\begin{outline}
\1 Racer API:
\2 Low level control:
\3 Angle rate setpoints: \texttt{moveByAngleRatesThrottle, moveByAngleRatesZ}
\3 Angle setpoints: \texttt{moveByRollPitchYawThrottle, moveByRollPitchYawZ} 
\3 Roll, pitch angles and yaw rate: \texttt{moveByRollPitchYawrateThrottle, \\ moveByRollPitchYawrateZ} 
\3 Set motor RPM\textsuperscript{\#}: \texttt{setMotorSpeeds} 
\3 Set low level controller gains: \texttt{setAngleRateControllerGains, \\ setAngleLevelControllerGains}

\2 Medium level control:
\3 Velocity setpoints: \texttt{moveByVelocity, moveByVelocityZ}
\3 Position setpoints: \texttt{moveToPosition, moveToZ, moveOnPath}
\3 Set medium level control gains: \texttt{setVelocityControllerGains, \\ setPositionControllerGains}

\2 High level - trajectory planning and tracking:
\3 Fit a spline through waypoints: \texttt{moveOnSpline},\\ \texttt{moveOnSplineVelConstraints} 
\3 Track a trajectory: \texttt{trackTrajectory}
\3 Set control gains for tracking: \texttt{setTrajectoryTrackerGains}

\1 Environment API:
\2 Race environment API:

\3 Load map: \texttt{simLoadlevel}
\3 Spawning and destruction of gates: \texttt{simSpawnObject, simDestroyObject}
\3 Gate size ground truth: \texttt{simGetNominalGateInner/OuterDimensions}

\2 Object API: (To facilitate dataset generation)

\3 Object pose and scale: \texttt{simSet/GetObjectPose, simSet/GetObjectScale}
\3 Object segmentation: \texttt{simSet/GetSegmentationObjectID}
\3 List all the objects in the scene: \texttt{simListSceneObjects}.
\3 Domain Randomization$^{\#}$: \texttt{simSetTexture}

\1 Race API:
\2 Starting and resetting of a race: \texttt{simStartRace, simResetRace}.
\2 Tracking of race progress: \texttt{simGetLastGatePassed}, \texttt{simIsRacerDisqualified}
\2 Get collision information: \texttt{simGetCollisionInfo}
\2 Disabling the generation of race log files: \texttt{simDisableRaceLog}.

\1 Utilities:
\2 Environment Geometry API$^{\#}$:
\3 Voxel grids and Signed Distance Field construction: \texttt{buildVoxelGrid/buildSDF} 
\3 Occupancy and signed distance for points: \texttt{isOccupied, getSDF}
\3 Signed distance field gradients: \texttt{getSDFGradient}
\3 Project to nearest free space: \texttt{getNearestFreePt} 

\2 Camera Model API$^{\#}$: 
\3 Change camera field of view: \texttt{camSetFoV} 
\3 Change camera distortion parameters: \texttt{camSetDistortionParams}
\3 Enable rolling shutter effect: \texttt{camEnableRollingShutter}

\2 Event camera API$^{\#}$:
\3 Generate event data: \texttt{generateEvents}
\3 Set event camera parameters (contrast threshold, noise etc.): \texttt{camSetEventParams}

\1  Visualization API$^{\#}$: 
\2  Suite of functions aimed at plotting for debug purposes: \texttt{simPlotPoints, \\ simPlotMarkers, simPlotString, simPlotTrajectory, simPlotTransform}

\end{outline}

\section{Race track complexity}
\label{appendix:trackcomplexity}

\begin{figure*}
	\centering
	\subfloat[Track complexity for planning and control tasks, measured in terms of instantaneous curvature. Note the different scales on X and Y axis.\label{fig:tracks_curvature}]{\includegraphics[width=\textwidth]{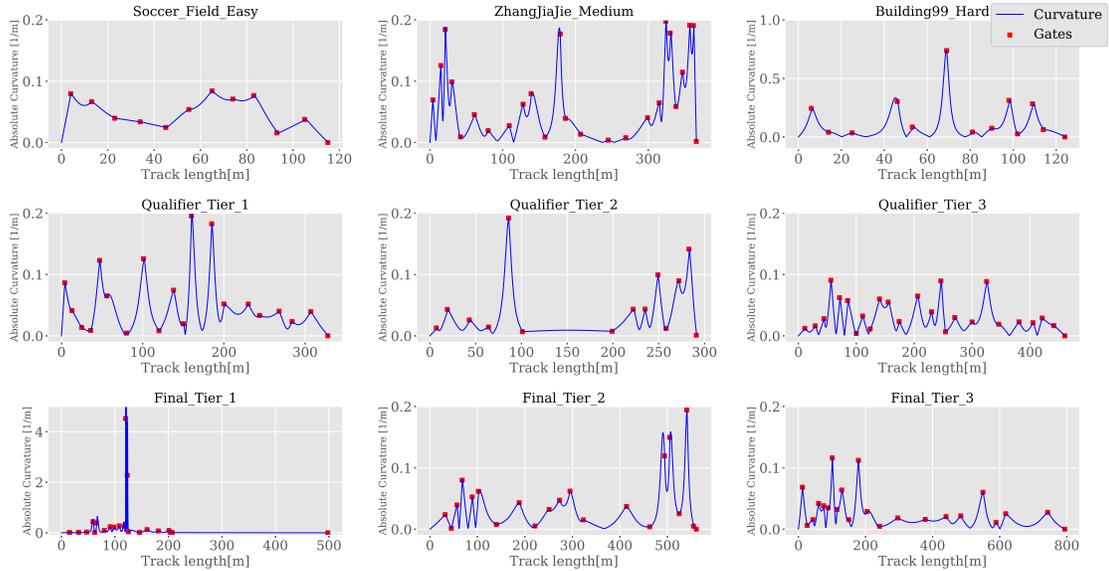}}
	\qquad
	\subfloat[Track complexity for perception tasks, measured in terms of next gate visibility in the image. Note that each track is of different length, and the X-axis is normalized\label{fig:tracks_gate_visibility}] {\includegraphics[width=\textwidth]{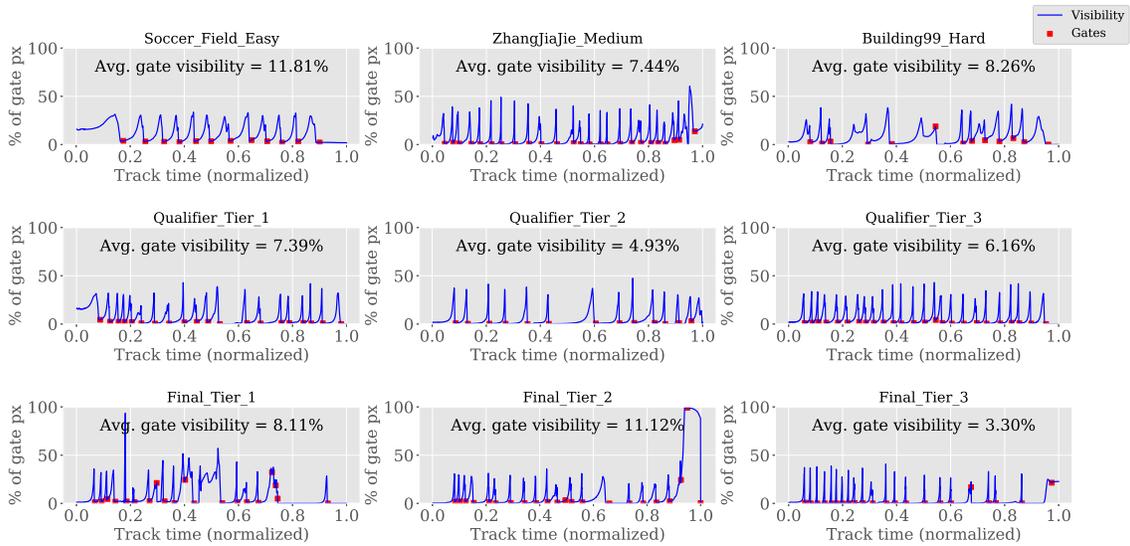}}
	\caption{Track Complexity}
	\label{fig:track_complexity}
\end{figure*}

The race track complexity metrics for all tracks are visualized in \autoref{fig:track_complexity}.
\autoref{fig:tracks_curvature} visualizes how the curvature changes along each race track, and \autoref{fig:tracks_gate_visibility} visualizes how the gate visibility (number of image pixels occupied by the gate divided by total number of pixels) varies along each track.

\section{Events and optical flow generation}
\label{appendix:eventsandopticalflow}

For simulating events from the RGB image stream of a camera, we loosely follow the framework presented in \cite{Rebecq18corl}. As an improvement to the above approach, we utilize Unreal Engine's graphics capabilities directly for improved performance. Pixel intensity differences are first computed through image subtraction, which is written as a shader that runs directly on the GPU (thus natively parallelized). This allows for calculation and rendering of pixel differences at near-realtime frame rates. This representation is then converted into simulated asynchronous event data (which is in process), again by utilizing compute shader nodes written in High Level Shading Language (HLSL) to ensure high performance and tight coupling between the renderer and the event generation. 

We also provide access to ground truth optical flow images, which can be used for learning motion models and pose predictions (\cite{fischer2015flownet}, \cite{maurer2018proflow}). In the simulator, we obtain optical flow information by accessing Unreal Engine's rendering pass, which also encodes screen space velocities for all pixels; information that is subsequently used also to compute motion blur. 

\newpage